\documentclass[runningheads]{llncs}
\usepackage[T1]{fontenc}
\usepackage{graphicx}
\usepackage{booktabs}
\usepackage[misc]{ifsym}

\usepackage{multirow}
\usepackage{xcolor,colortbl}

\usepackage{amsmath}
\usepackage{amssymb}
\usepackage[ruled,linesnumbered]{algorithm2e} % 更现代的算法包
\usepackage{algpseudocode}

\usepackage{mwe}
\usepackage[misc]{ifsym}
\begin{document}

\title{LeAP: Learnable Adaptive Permutation for Feature Selection in Heterogeneous and  Sparse Recommender Systems}

\titlerunning{Learnable Adaptive Permutation}
% If the full title of your paper is short enough to also fit in the running head, you can omit the abbreviated paper title here. You can check as follows: if you comment out the \titlerunning line, something will appear in the header of all odd-numbered pages of your PDF from page 3 onward. This something is either the full title (in which case all is well), or the error message "Title Suppressed Due to Excessive Length". If this error message appears, you're going to want to provide an abbreviated title within the \titlerunning command, because if you won't do it, Springer will do it for you.

%N.B.: Author information (both in the \author{} and \authorrunning{} command) should only be present in the Camera-Ready Version of your paper. The version that you initially submit for review, ought to be double-blind. So, when initially submitting your paper, use:
%\author{Author information scrubbed for double-blind reviewing}

% \author{Andr\'e Lauren Benjamin\inst{1} \and
% Calvin Cordozar Broadus Jr.\inst{2,3} \corr \and
% Antwan Andr\'e Patton\inst{1}}

% % 匿名作者
% \author{Anonymous Author(s)}

% \authorrunning{Anonymous Author(s)}
% 最终投稿
% ======= 作者姓名与关联设置 =======
\author{Yihong Huang \and
Chen Chu \Letter \thanks{Corresponding author.}  \and
Fei Chen \and
Yu Lin \and
Ruiduan Li \and
Zhihao Li}

% ======= 运行页眉作者缩写 =======
\authorrunning{Y. Huang et al.}

% ======= 机构与邮箱设置 =======
\institute{Bilibili Inc., Shanghai, China \\
\email{hyh957947142@gmail.com, chuchen.blueblues@gmail.com, chenfei03@bilibili.com, linyu03@bilibili.com,
ruidli1992@gmail.com,
zhihao.lee@foxmail.com
}}

% ruidli1992\}@gmail.com}, \\
% \email{chenfei03@bilibili.com}, \email{linyu03@bilibili.com}, \email{zhihao.lee@foxmail.com}}

% ======= 目录页设置（根据邮件要求新增） =======
\tocauthor{Yihong Huang, Chen Chu, Fei Chen, Yu Lin, Ruiduan Li, Zhihao Li}
\toctitle{LeAP: Learnable Adaptive Permutation for Feature Selection in Heterogeneous and Sparse Recommender Systems}
% \email{\{hyh957947142, chuchen.blueblues, ruidli1992\}@gmail.com}, \\
% \email{chenfei03@bilibili.com}, \email{linyu03@bilibili.com}, \email{zhihao.lee@foxmail.com}}

% You may leave out the orcidID information, if you want to.
% Use \corr to indicate the corresponding author. Note the spacing around the \corr command. Only one author can be the corresponding author.

%N.B.: comment out the \authorrunning{} command for the double-blind version of your paper submitted for review. Later, if your paper is accepted, use the command for the Camera-Ready Version.
% \authorrunning{A.L. Benjamin et al.}
% First names are abbreviated in the running head.
% If there is one author, write 'A.L. Benjamin'.
% If there are two authors, write 'A.L. Benjamin and C.C. Broadus Jr.'
% If there are more than two authors, '[...] et al.' is used.

% \institute{Fictional Southern University, Savannah GA 31404, USA \email{\{a.l.benjamin,a.a.patton\}@fsu.fake}
% \and
% Fictional West Coast University, Long Beach CA 90840, USA \email{ccb@fwcu.fake}
% \and
% Secondary European Affiliation, Tiergartenstr. 17, 69121 Heidelberg, Germany
% \email{lncs@springer.com}}

\maketitle              % typeset the header of the contribution

\begin{abstract}
Modern industrial recommender systems rely on thousands of heterogeneous features—ranging from low-dimensional scalars (e.g., statistical value) to high-dimensional embeddings (e.g., user-id embeddings, MLP representations)—to achieve high-precision predictions. Given the immense computational costs associated with training, efficient feature selection is critical. However, existing methods encounter three primary bottlenecks: (1) they typically assume uniform feature dimensions or require costly mapping to a fixed size; (2) they struggle with extreme sparsity, where the majority of features (e.g., 99\%+) remain at default values; and (3) traditional permutation-based approaches are computationally prohibitive in large-scale settings.

To address these challenges, we propose LeAP (Learnable Adaptive Permutation), a novel, model-agnostic plug-in module for feature selection. LeAP transforms the inefficient random permutation process into a learnable mechanism, significantly accelerating the evaluation of feature importance. In addition, we introduce an adaptive regularization strategy tailored for heterogeneous dimensions and extreme sparsity, enabling superior feature importance ranking results across asymmetric input spaces. Experiments on four public recommendation datasets demonstrate that LeAP achieves state-of-the-art performance. Furthermore, LeAP has been deployed in a large-scale industrial search ranking model with over a billion daily requests and a 2TB model parameter scale. In this real-world scenario involving 12,000+ total  feature dimensions, LeAP successfully identified and removed over 3,600 redundant dimensions without performance degradation, which is 2 to 10 times the ability of compared baseline methods. 

\end{abstract}

\section{Introduction}

In modern industrial search \cite{yahoo-search}, advertising \cite{online-ad}, and recommendation systems \cite{recommendation-survery,netflix}, the predictive accuracy of deep learning models heavily relies on massive and complex feature engineering. To accurately capture dynamic user intents and content representations, production models typically process thousands of features. These features exhibit strong \textbf{dimensional heterogeneity} and type diversity \cite{gu2020deep,youtubeDRS}: ranging from 1-dimensional continuous scalars (e.g., real-time statistics, numerical features) to vectors spanning tens to hundreds of dimensions (such as user behavior embeddings \cite{din} or MLP-compressed latent representations). Such complexity in the feature space pushes industrial model sizes to the terabyte (TB) level \cite{rankmixer}, incurring staggering computational and storage costs for both training and online serving.

\textbf{\textit{Challenge}.} To alleviate the computational burden and eliminate outdated or redundant information, feature selection \cite{zheng2023automl,bolon2013review} has become a critical component of model optimization. However, existing feature selection methods encounter significant bottlenecks when deployed in these complex industrial scenarios. First, \textbf{most existing approaches are oblivious to the dimensional discrepancy of heterogeneous features.} They often assume homogeneous feature dimensions or enforce mapping all features into a unified dimension space \cite{autofield}. This not only distorts the original information density but also makes it profoundly difficult to fairly evaluate the true contributions of a 1D feature against a 256D feature under the same metric. Second, \textbf{existing methods are highly unfriendly to sparse features.} Because sparse features remain at default states in the vast majority of instances, existing selection mechanisms \cite{deeplasso} often misinterpret their "low frequency of activation" as "useless noise," penalizing or discarding them heavily, thereby severely impairing the representation of long-tail personalized signals.

Finally, although methods based on Permutation Feature Importance \cite{permutation,shark} offer high interpretability and model-agnostic capabilities in evaluating true feature contributions, they suffer from \textbf{prohibitive computational inefficiency}. Traditional permutation methods require sequentially shuffling each feature in isolation and re-executing the forward pass. For ultra-large-scale models with thousands of features, this one-variable-at-a-time permutation paradigm results in an unbearable computational disaster, making it entirely infeasible to integrate into the daily iterations of industrial models that process billion-level requests.

\textbf{\textit{Method}.} To break through these dilemmas, we propose \textbf{LeAP (Learnable Adaptive Permutation)}, a model-agnostic, plug-in feature selection module explicitly tailored for large-scale industrial models. The core breakthrough of LeAP lies in transforming the traditionally discrete and computationally prohibitive permutation evaluation into an \textbf{end-to-end learnable mechanism}. During a single forward pass of model training, LeAP employs a dynamic gating network to learn the fusion weights between the original features and their batch-wise permuted noise counterparts. This design thoroughly circumvents the performance bottleneck of repetitive inference required by traditional permutation methods, drastically reducing the computational complexity from $O(N \times \text{model overhead})$ to $O(1 \times \text{model overhead})$. Consequently, LeAP can be effortlessly embedded into daily training pipelines handling tens of billions of samples.

More crucially, to tackle the notoriously difficult problems of feature heterogeneity and sparsity in industrial settings, LeAP introduces an innovative Adaptive Regularization based on Permutation Divergence. Specifically, LeAP calculates the $L_2$-norm difference between the original and permuted inputs for each feature in real-time. By applying Exponential Moving Average (EMA) to construct stable statistical bounds, this divergence serves as an adaptive base penalty for the feature's gating weight.

This mechanism exhibits two remarkably elegant desirable properties:

(1) \textbf{Dimensionally Adaptive}: For wide features such as 128D or 256D embeddings, the spatial distance before and after permutation is inherently larger than that of 1D scalars. LeAP adaptively imposes a stronger penalty on high-dimensional features, fundamentally offsetting the unfair advantage they often exploit to "hide" from traditional uniform regularization.

(2) \textbf{Sparsity-Aware}: For extremely sparse features where 99\% of instances hold default values, the permuted values are highly likely to remain defaults. Consequently, their $L_2$ difference is minimal, and the penalty force shrinks accordingly. This establishes a vital criterion for industrial feature selection: we should not severely penalize a feature simply because it is sparse; it should be discarded only if it is genuinely useless to the predictive objective.

Empowered by this adaptive penalty, LeAP not only outputs highly interpretable feature importance scores (reflecting the model's sensitivity to information loss) but also drives the gating weights to naturally polarize. This polarization implies that redundant features can be directly pruned via straightforward simple thresholding. It eliminates the need for the resource-intensive model retraining typically required by prior methods, allowing for direct online serving with minimal fine-tuning.

\textbf{\textit{Experimental Results.}} 
To comprehensively evaluate the effectiveness of LeAP, we conducted extensive experiments on both public datasets and real-world industrial production environments. On four public recommender system datasets, LeAP significantly outperforms existing mainstream and state-of-the-art feature selection methods, consistently achieving SOTA predictive performance across varying feature pruning ratios.

More importantly, LeAP has been successfully deployed in an industrial search ranking model with billions of daily requests. In actual production, the model's heterogeneous input features, spanning over 12,000 dimensions, create a severe CPU-to-GPU memory bandwidth bottleneck. Furthermore, since all existing online features have undergone rigorous business validation, performing secondary pruning among these high-value features poses a tremendous challenge. Nevertheless, LeAP exhibits remarkable precision and robustness: in the evaluation of feature pruning ROI (Return on Investment), traditional methods suffer significant performance degradation when pruning fewer than 600 dimensions; in contrast, LeAP successfully identifies and eliminates over 30\% (3,600+) of the redundant dimensions.  It successfully resolves the system bandwidth bottleneck while achieving zero degradation (Zero Diff) in core online business metrics.

In summary, the primary contributions of this paper are as follows:
\begin{itemize}
    \item \textbf{Engineering Architecture Innovation}: We propose LeAP, a model-agnostic feature selection plug-in. It reconstructs the inefficient, discrete feature permutation process into an $O(1)$ end-to-end learnable mechanism, thoroughly breaking the computational bottleneck of traditional permutation-based feature evaluation.
    
    \item \textbf{Algorithmic Theory Breakthrough}: We design an  \textbf{Adaptive Regularization based on Permutation Divergence}. This mechanism mathematically overcomes the evaluation biases introduced by heterogeneous dimensions (e.g., 1D vs. 256D) and extremely sparse features, enabling genuinely fair feature selection.
    \item \textbf{Substantial Industrial Deployment Gains}: LeAP not only achieves SOTA results on public datasets but also successfully prunes over 30\% of feature dimensions in a 2TB-scale real-world search model with zero loss in business metrics. This provides a highly valuable industrial practice paradigm for feature pruning in ultra-large-scale recommender systems.
\end{itemize}

Our code is available at \url{https://github.com/goldenNormal/LeAP}.

\section{Related Work}

\subsection{Differentiable Mask-based Feature Selection}

Feature selection (FS) is indispensable for mitigating the storage and computational overhead in industrial recommender systems. While traditional statistical and tree-based methods \cite{lasso,rf,lightgbm,xgboost} offer interpretability, they are ill-suited for high-dimensional embeddings and cannot be optimized jointly with deep serving models.

Recent paradigms have shifted towards differentiable mask-based methods \cite{autofield,lpfs,sfs,optfs}, which integrate learnable gates directly into the model architecture. For instance, AutoField \cite{autofield} employs Gumbel-Softmax for feature gating, and LPFS \cite{lpfs} utilizes a smoothed $L_0$ regularization. Despite their success on standard benchmarks, these methods implicitly assume a \textbf{homogeneous feature space} (e.g., assuming all fields are embedded into the same dimension $D$). When applied to heterogeneous industrial input layers—where dense 1D scalars are concatenated with sparse high-dimensional embeddings—these methods typically apply uniform regularization scales. Consequently, high-dimensional representations generate disproportionately larger gradients and variance compared to 1D scalars. This imbalance leads to a biased optimization process where critical low-dimensional signals are heavily penalized and prematurely pruned.

While Dimension Selection (DS) \cite{autodim,dimreg,IRazor} attempts to assign varying dimensions to different fields, these approaches typically require initializing all fields into a uniform "super-net" embedding space before pruning. Forcing native 1D statistical features into high-dimensional spaces introduces parameter redundancy and information dilution. In contrast, \textbf{LeAP operates directly on the native heterogeneous concatenation layer}, ensuring a balanced importance assessment without distorting the original feature representations.

\subsection{Permutation-based Feature Importance}

Another prevalent paradigm assesses feature importance via Permutation \cite{permutation,shark}. By measuring the prediction degradation when a specific feature is replaced by random noise across instances, Permutation provides model-agnostic evaluation and clear interpretability.

However, applying permutation-based methods in industrial settings reveals two critical limitations. First, they suffer from \textbf{substantial computational overhead}. Standard Permutation and its Taylor-approximated variants, such as SHARK \cite{shark}, require $O(N)$ repetitive inference passes and iterative "train-prune-train" cycles. 
Second, \textbf{permutation evaluation is inherently biased against extremely sparse features}. For features where the vast majority of instances (e.g., >99\%) hold default values, shuffling predominantly exchanges default values with other default values. Consequently, the observed performance drop is minimal, causing standard permutation methods to underestimate the true importance of these features and mistakenly prune critical long-tail signals.

\textbf{To address the limitations of both aforementioned paradigms, our proposed LeAP serves as a unified plug-in module.} It synergizes the $O(1)$ efficiency of differentiable mask-based methods with the interpretability of permutation-based approaches. By shifting from standard discrete permutation evaluation to a learnable permutation divergence mechanism, LeAP effectively avoids the pitfalls of both paradigms. It provides a cohesive solution to the challenges of heterogeneous dimensions and extreme sparsity, requiring minimal modifications to the existing model architecture.

\section{Methodology}

\begin{figure}[t]
    \centering
\includegraphics[width=\linewidth]{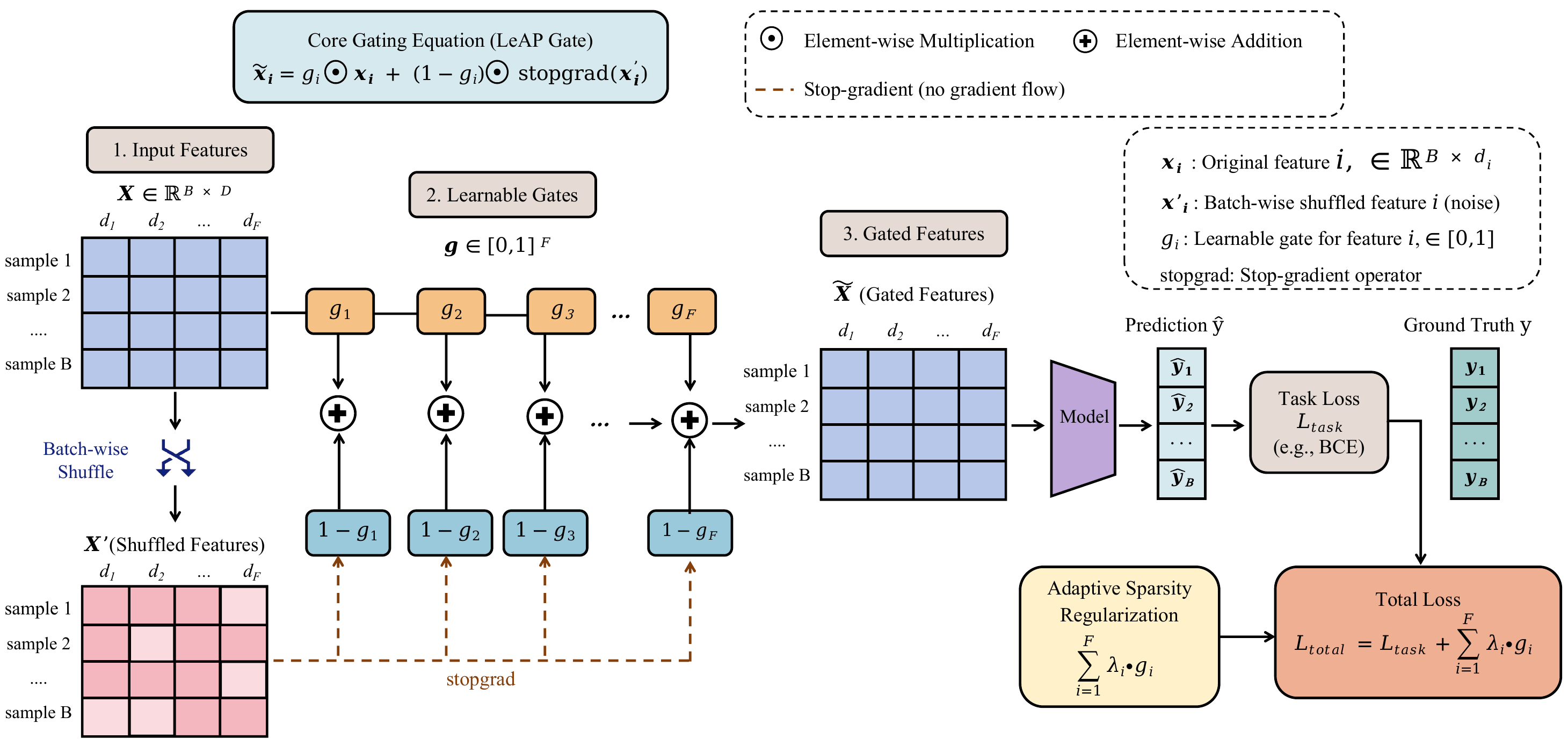}
    \caption{The overview of LeAP Module.}
    \label{fig:leap-module}
\end{figure}

In this section, we detail the overall architecture and mathematical foundations of \textbf{LeAP} (Learnable Adaptive Permutation) module, as shown in Figure \ref{fig:leap-module}. We first introduce the transformation of traditional discrete permutation into an end-to-end learnable gating mechanism in Section 3.1. Subsequently, in Section 3.2, we elaborate on our core innovation for tackling dimensional heterogeneity and extreme sparsity: Adaptive Regularization via Permutation Divergence. Finally, we provide a theoretical analysis of how this mechanism drives gating polarization (Section 3.3) and discuss the deployment paradigm in real-world production environments (Section 3.4).

\subsection{Learnable Permutation Module}

In deep recommender systems, model inputs typically consist of $F$ feature fields. Due to the diversity of feature types, these features exhibit severe dimensional heterogeneity when mapped into the latent space. Let the representation of the $i$-th feature be denoted as $\mathbf{x}_i \in \mathbb{R}^{d_i}$, where $d_i$ varies drastically (e.g., $d_i=1$ for scalar statistical features, whereas $d_i=256$ for user behavior sequence features).

To evaluate feature importance without requiring multiple forward passes, LeAP introduces a batch-wise learnable permutation mechanism. At each training step, given a mini-batch of size $B$, we first shuffle the $i$-th feature $\mathbf{x}_i$ along the batch dimension to obtain its noise counterpart $\mathbf{x}_i'$. This shuffling operation not only breaks the correlation between the feature and the target label but also perfectly preserves the feature's original marginal distribution and sparsity pattern. Algorithm \ref{alg:efficient_shuffle} illustrates an implementation.

Next, we introduce a feature-specific learnable gating variable $g_i \in (0, 1)$ to control the information fusion between the original and shuffled features. To ensure end-to-end differentiability, we parameterize it using a temperature-scaled Sigmoid function:

$$g_i = \sigma \left( \frac{\theta_i}{\tau} \right) = \frac{1}{1 + \exp(-\theta_i / \tau)}$$
where $\theta_i \in \mathbb{R}$ is the learnable parameter tied to the $i$-th feature, and $\tau$ is the temperature hyperparameter controlling the smoothness.

The fused output feature $\tilde{\mathbf{x}}_i$ can be expressed as a convex combination of the original and shuffled features $\mathbf{x}_i'$. To ensure the gradients only update the gating weights and do not propagate through the noise to disrupt upstream layers, we apply a stop-gradient operation $\text{sg}(\cdot)$ to the shuffled feature:

$$\tilde{\mathbf{x}}_i = g_i \cdot \mathbf{x}_i + (1 - g_i) \cdot \text{sg}(\mathbf{x}_i')$$

\textbf{Interpretation:} Under this formulation, $g_i$ intuitively reflects the model's reliance on the original feature. As $g_i \to 1$, the model fully utilizes the genuine feature; as $g_i \to 0$, the model tends to rely on the permuted pure noise. If a feature is redundant, replacing it with noise will not degrade the model's predictive performance.

\begin{algorithm}[h]
\SetAlgoLined
\KwIn{$\mathbf{X} \in \mathbb{R}^{B \times D}$: Input tensor within a batch, $D = \sum d_i$\\
      $\mathcal{F} = \{(s_1, e_1), ..., (s_F, e_F)\}$: Feature dimension ranges}
\KwOut{${\mathbf{X}'} \in \mathbb{R}^{B \times D}$: Shuffled tensor}
${\mathbf{X}'} \gets \mathbf{0}^{B \times D}$ \tcp*{Initialize output tensor}
\For{$(s_i, e_i) \in \mathcal{F}$}{
    $\pi_i \gets \text{RandomPermutation}(B)$ \\
    ${\mathbf{X}'}_{:,s_i:e_i} \gets \mathbf{X}_{:,s_i:e_i}[\pi_i,:]$ \tcp*{Apply shuffling}
}
\Return{${\mathbf{X}'}$}
\caption{Batch Shuffling}
\label{alg:efficient_shuffle}
\end{algorithm}

\subsection{Adaptive Regularization via Permutation Divergence}

In standard gated feature selection methods \cite{deeplasso,fea-l1-reg,dimreg,lpfs}, a uniform penalty (e.g., $\lambda \sum g_i$) is typically applied to all gating variables to encourage sparsity. However, when applied directly to the heterogeneous and sparse feature layers in industrial settings, this uniform penalty induces severe evaluation unfairness. High-dimensional features, due to their larger parameter volume, naturally generate greater gradient fluctuations than 1D features. Conversely, extremely sparse features (which are predominantly zero) produce weak gradients, making them highly susceptible to being mistakenly pruned by a uniform regularization term.

To address this critical limitation, LeAP discards the static uniform penalty and innovatively proposes a \textbf{data-aware adaptive regularization mechanism}. We argue that the penalization strength on a feature should not be based on its predefined dimensionality, but rather on the magnitude of information perturbation induced by its permutation.

Specifically, within each mini-batch, we calculate the $L_2$-norm difference (Permutation Divergence) between the original feature $\mathbf{x}_i$ and its shuffled version $\mathbf{x}_i'$:

$$\Delta_i = \frac{1}{B} \sum_{b=1}^{B} \| \mathbf{x}_i^{(b)} - \mathbf{x}_i'^{(b)} \|_2$$

To mitigate data variance across batches and establish a stable statistical bound, we employ an Exponential Moving Average (EMA) to smooth this divergence throughout the training process:

$$\bar{\Delta}_i^{(t)} = \beta \cdot \bar{\Delta}_i^{(t-1)} + (1 - \beta) \cdot \Delta_i^{(t)}$$
where $\beta \in [0, 1)$ is the momentum coefficient, and $t$ denotes the training step. Additionally, to identify constant features, we apply appropriate clipping to $\bar{\Delta}_i$.

Based on this stable permutation divergence, we formulate the adaptive regularization weight $\lambda_i$ for feature $i$:

$$\lambda_i = \alpha \cdot \bar{\Delta}_i$$
where $\alpha$ is a global hyperparameter controlling the overall feature sparsity. Finally, the end-to-end optimization objective (Total Loss) of LeAP is defined as the sum of the task loss and the adaptive regularization term:

$$\mathcal{L}_{total} = \mathcal{L}_{task} + \sum_{i=1}^{F} \lambda_i \cdot g_i$$

This highly streamlined mathematical design simultaneously resolves two major industrial pain points:

1. \textbf{Inherent Dimensional Normalization}: The $L_2$ divergence $\bar{\Delta}_i$ of a shuffled 128D embedding is intrinsically much larger than that of a 1D scalar. Consequently, high-dimensional features automatically incur a larger penalty $\lambda_i$. This adaptive reweighting fundamentally counteracts the tendency of ineffective parameters in high-dimensional spaces to evade penalization, ensuring fair competition across varying dimensions.

2. \textbf{Perfect Tolerance for Sparsity}: For extremely sparse features where 99\% of instances hold default values, the shuffling process highly likely swaps default values with other default values. In this scenario, the $L_2$ divergence $\bar{\Delta}_i \to 0$, resulting in a negligible regularization penalty $\lambda_i$. This guarantees that sparse features are not mistakenly pruned simply due to their "low frequency of occurrence"; they are gradually phased out only if they contribute genuinely nothing to the predictive objective $\mathcal{L}_{task}$.

\subsection{Theoretical Analysis}

Existing gating methods \cite{autofield,sfs} often produce ambiguous feature importance scores. In contrast, the core advantage of LeAP lies in its ability to drive gating scores to naturally polarize towards 0 or 1. We provide a theoretical justification for this phenomenon from the perspectives of gradient decoupling and convex optimization.

\noindent\textbf{1. Gradient Decoupling and Scale Alignment.} 
Recall the fusion formulation $\tilde{\mathbf{x}}_i = g_i \cdot \mathbf{x}_i + (1 - g_i) \cdot \mathbf{x}_i'$. The gradient of the expected task loss $J(g_i) = \mathbb{E}[\mathcal{L}_{\text{task}}]$ with respect to the gate $g_i$ can be decoupled via the chain rule:

$$\frac{\partial J(g_i)}{\partial g_i} = \mathbb{E} \left[ \underbrace{\left( \frac{\partial \mathcal{L}_{\text{task}}}{\partial \tilde{\mathbf{x}}_i} \right)^T}_{\text{Sensitivity } \mathbf{S}_i} \cdot \underbrace{(\mathbf{x}_i - \mathbf{x}_i')}_{\text{Permutation Divergence}} \right]$$

This equation reveals that the magnitude of the task gradient is naturally proportional to the permutation divergence $\|\mathbf{x}_i - \mathbf{x}_i'\|_2$. Therefore, by setting the adaptive penalty to $\lambda_i = \alpha \cdot \bar{\Delta}_i$, LeAP mathematically achieves an effective scale alignment between the regularization term and the task gradient. This alignment effectively resolves the evaluation bias that high-dimensional and extremely sparse features encounter under uniform penalties.

\noindent\textbf{2.  Polarized Gate Guarantee.} 
Based on the aligned total objective function $\mathcal{J}_{\text{total}}(g_i) = J(g_i) + \lambda_i g_i$, the total gradient is given by $\frac{\partial \mathcal{J}_{\text{total}}}{\partial g_i} = \frac{\partial J(g_i)}{\partial g_i} + \lambda_i$. We demonstrate that genuinely useful features are guaranteed to polarize towards 1.

\textbf{Theorem 1.} \textit{Assume the expected task loss $J(g_i)$ is convex. Let $\Delta J = J(0) - J(1)$ denote the loss increment after shuffling the feature (i.e., the signal strength). If $\Delta J > \lambda_i$, the total gradient is negative, and the gate $g_i$ will be driven towards 1.}

\textit{Proof.} Based on the first-order properties of convex functions, the function curve lies below the chord connecting its endpoints, meaning $J(g_i) \le g_i J(1) + (1 - g_i) J(0)$. Rearranging this inequality and taking the partial derivative yields an upper bound for the task gradient:

$$\frac{\partial J(g_i)}{\partial g_i} \le J(1) - J(0) = -\Delta J$$

Substituting this into the total gradient formulation gives:

$$\frac{\partial \mathcal{J}_{\text{total}}}{\partial g_i} \le -\Delta J + \lambda_i$$

Given the assumption that the valid feature's signal strength satisfies $\Delta J > \lambda_i$, we have $\frac{\partial \mathcal{J}_{\text{total}}}{\partial g_i} < 0$. The optimizer will thus continuously increase $g_i$ until it reaches the upper bound of 1. 

\textbf{Corollary.} \textit{Conversely, if a feature is redundant, shuffling it does not impact the prediction, resulting in $\Delta J \approx 0$. The total gradient then degenerates to $\approx \lambda_i > 0$. The regularization term dominates the optimization, rapidly suppressing the gate $g_i$ towards 0.}

\subsection{Practical Deployment in Industrial Systems}

To meet the stringent iteration and latency requirements of large-scale industrial recommender systems, LeAP adopts a highly decoupled plug-in deployment paradigm. In real-world production environments, the pipeline consists of three streamlined steps:

\noindent\textbf{1. Plug-in \& Evaluation.} 
Extract the online  model and insert the LeAP module immediately after the feature concatenation layer. The model then iterates normally using the online data stream, accumulating the EMA statistics and updating the gating variables.

\noindent\textbf{2. Hard Pruning via Thresholding.} 
Thanks to the polarization property, the converged gates $g_i$ exhibit strong discriminability. Once the evaluation phase concludes, feature pruning can be executed directly using two strategies:
\begin{itemize}
    \item \textbf{Absolute Thresholding:} Set a fixed threshold (e.g., $\tau = 0.5$) and directly discard all features where $g_i < \tau$.
    \item \textbf{Proportional Ranking:} When facing strict system bandwidth reduction constraints, sort the features in descending order of their $g_i$ scores and retain only the Top-$K$ proportion.
\end{itemize}

\noindent\textbf{3. Fine-tuning \& Serving.} 
After removing the redundant features and the LeAP module itself, the model only requires fine-tuning on the small fraction of training data  to convergence.

% \section{Experiments}
\section{Experiments}

In this section, we comprehensively evaluate the effectiveness and engineering value of LeAP through extensive offline evaluations on public datasets and a large-scale industrial deployment. Our experiments are designed to answer the following core research questions:
\begin{itemize}
    \item \textbf{RQ1 (Basic Efficacy):} On homogeneously dimensioned public datasets, can the fundamental mechanism of LeAP outperform existing state-of-the-art (SOTA) feature selection methods?
    \item \textbf{RQ2 (Mechanism Polarization):} Does LeAP genuinely generate highly polarized gating scores with clear physical interpretations, as theoretically proven?
    \item \textbf{RQ3 (Industrial Deployment \& Ablation):} In real-world industrial scenarios characterized by extreme dimensional heterogeneity and sparsity, what is the Return on Investment (ROI) of LeAP? 
\end{itemize}
Our code is available at \url{https://github.com/goldenNormal/LeAP}.

\subsection{Experimental Setup}

\noindent\textbf{Datasets:}
\begin{itemize}
    \item \textbf{Public Datasets:} We utilize four widely adopted benchmark datasets \cite{ERASE} in the recommender system domain: Avazu\footnote{\url{https://www.kaggle.com/competitions/avazu-ctr-prediction}}, Criteo\footnote{\url{https://ailab.criteo.com/ressources/}}, MovieLens-1M (ML-1M)\footnote{\url{https://grouplens.org/datasets/movielens/1m/}}, and AliCCP\footnote{\url{https://tianchi.aliyun.com/dataset/408}}. The feature dimensions of these datasets have been set to a uniform size.
    \begin{table}[h]
    \centering
    \caption{Dataset statistics.}
    \label{tab:dataset}
    \small
    \setlength{\tabcolsep}{4pt}
    \begin{tabular}{lcccc}
        \toprule
        \textbf{Dataset} & \textbf{Avazu} & \textbf{Criteo} & \textbf{ML-1M} & \textbf{AliCCP} \\
        \midrule
        \textbf{Samples} & 40,428,967 & 45,850,617 & 1,000,209 & 85,316,519 \\
        \textbf{Label} & Click & Click & Rating (1-5) & Click \\
        \textbf{Fields} & 23 & 39 & 9 & 23 \\
        \bottomrule
    \end{tabular}
\end{table}

    \item \textbf{Industrial Dataset:} This dataset is collected from real-world search, impression,  click and interaction logs of a long-video platform with over a billion daily requests. The model incorporates more than 500 validated feature fields, resulting in a concatenated representation exceeding 12,000 dimensions. The feature dimensions vary drastically, ranging from 1D (e.g., statistical features) to 256D (e.g., aggregated behavioral sequence embeddings), and include a substantial number of extremely sparse features.
\end{itemize}

\noindent\textbf{Baselines:}
We select seven representative feature selection methods as baselines, categorized as follows:
(1) \textbf{Heuristic:} Lasso \cite{lasso}, Random Forest \cite{rf} (RF), XGBoost \cite{xgboost}.
(2) \textbf{Mask-based:} AutoField \cite{autofield}, LPFS \cite{lpfs}, SFS \cite{sfs}.
(3) \textbf{Permutation-based:} SHARK \cite{shark} (an industry-recognized efficient variant of permutation). 

Since the public datasets possess homogeneous feature dimensions, we deploy the base version of LeAP (without adaptive regularization) in Section 4.2 to solely validate the superiority of its native learnable permutation mechanism.

\noindent\textbf{Evaluation Metrics:}
For public datasets, we employ AUC and the cross-dataset aggregated metric, Normalized AUC \cite{S_auc} ($S_{\text{AUC}}$):
\begin{equation}
S_\text{AUC}(A) = \frac{1}{|\Gamma|}\sum_{\mathcal{D} \in \Gamma} \frac{\text{AUC}(A,\mathcal{D})}{\max_{A' \in \mathcal{A}} \text{AUC}(A',\mathcal{D})}
\label{eq:s_auc}
\end{equation}

For offline evaluation in the industrial scenario, we use Group AUC for CTR prediction to reflect core predictive capability. For online deployment, we conduct strict A/B testing to observe core business metrics (e.g., video views, interactions and watch time).

\subsection{Offline Evaluation on Public Datasets}

Following the standard "Search-Retrain" two-stage evaluation protocol \cite{ERASE}, we test all methods using WideDeep \cite{widedeep} as the backbone network. Table \ref{tab:fs_widedeep} presents the retraining performance at target Feature Retention (FR) ratios of 50\% and 25\%.

\begin{table}[t]
\centering
\small % 纵向排布后空间变大，可以用 small 或 normalsize
\caption{Feature Selection and Retrain AUC Results. Best results are highlighted in \textcolor{red}{\textbf{red}}.  On ML-1M, all top-performing methods converged to the same feature subset, resulting in identical retrained AUC.}
\label{tab:fs_widedeep}
\begin{tabular}{l|l|cccc|c}
\toprule
\textbf{Ratio (FR)} & \textbf{Method} & \textbf{Criteo} & \textbf{Avazu} & \textbf{AliCCP} & \textbf{ML-1M} & $\mathbf{S_{\text{AUC}}}$ \\
\midrule
\multirow{9}{*}{\textbf{50\%}} 
 & \textit{No\_Select} & 0.8014 & 0.7882 & 0.6598 & 0.7950 & 0.9954 \\
 \cmidrule{2-7}
 & Lasso & 0.7462 & 0.7088 & 0.6054 & 0.6484 & 0.8872 \\
 & XGBoost & 0.7682 & 0.7412 & 0.6499 & \textcolor{red}{\textbf{0.8097}} & 0.9710 \\
 & RF & 0.7921 & 0.7869 & 0.6549 & 0.7924 & 0.9895 \\
 \cmidrule{2-7}
 & SHARK & 0.7974 & \textcolor{red}{\textbf{0.7870}} & 0.6582 & \textcolor{red}{\textbf{0.8097}} & 0.9977 \\
 \cmidrule{2-7}
 & LPFS & 0.7944 & 0.7728 & 0.6554 & \textcolor{red}{\textbf{0.8097}} & 0.9913 \\
 & SFS & 0.7969 & 0.7827 & 0.6579 & 0.7946 & 0.9915 \\
 & AutoField & 0.7974 & 0.7869 & 0.6567 & 0.8077 & 0.9965 \\
 \cmidrule{2-7}
 & \textbf{LeAP (Ours)} & \textcolor{red}{\textbf{0.7984}} & \textcolor{red}{\textbf{0.7870}} & \textcolor{red}{\textbf{0.6583}} & \textcolor{red}{\textbf{0.8097}} & \textcolor{red}{\textbf{0.9981}} \\
\midrule
\midrule
\multirow{9}{*}{\textbf{25\%}} 
 & \textit{No\_Select} & 0.8014 & 0.7882 & 0.6598 & 0.7950 & 0.9960 \\
 \cmidrule{2-7}
 & Lasso & 0.7003 & 0.6016 & 0.5804 & 0.5286 & 0.7928 \\
 & XGBoost & 0.7157 & 0.7091 & 0.5869 & 0.7338 & 0.8977 \\
 & RF & 0.7626 & 0.7637 & 0.6032 & 0.6946 & 0.9237 \\
 \cmidrule{2-7}
 & SHARK & 0.7717 & \textcolor{red}{\textbf{0.7723}} & 0.6461 & \textcolor{red}{\textbf{0.8078}} & 0.9805 \\
 \cmidrule{2-7}
 & LPFS & 0.7609 & 0.7714 & 0.6423 & \textcolor{red}{\textbf{0.8078}} & 0.9754 \\
 & SFS & 0.7569 & 0.7685 & 0.6452 & \textcolor{red}{\textbf{0.8078}} & 0.9733 \\
 & AutoField & 0.7738 & 0.7692 & 0.6485 & 0.8068 & 0.9808 \\
 \cmidrule{2-7}
 & \textbf{LeAP (Ours)} & \textcolor{red}{\textbf{0.7841}} & \textcolor{red}{\textbf{0.7723}} & \textcolor{red}{\textbf{0.6489}} & \textcolor{red}{\textbf{0.8078}} & \textcolor{red}{\textbf{0.9854}} \\
\bottomrule
\end{tabular}
\end{table}

\noindent\textbf{1. Overall Dominance:} Under both high compression rates of 50\% and 25\%, LeAP consistently achieves the highest $S_{\text{AUC}}$, which demonstrates its effectiveness in feature selection.

\noindent\textbf{2. Overcoming the "Weight Equals Importance" Assumption (Vs. Mask-based):} Methods like AutoField and LPFS operate on the assumption that "smaller gating weights imply less important features." However, in deep non-linear networks, features with small weights can trigger substantial "butterfly effects." LeAP discards weight magnitude in favor of measuring sensitivity to information loss. Consequently, it outperforms all mask-based methods by a significant margin on the complex Criteo and AliCCP datasets.

\noindent\textbf{3. Joint Optimization vs. Greedy Search (Vs. Permutation):} While both LeAP and SHARK are sensitivity-based methods, LeAP exhibits slightly superior performance. This is because SHARK employs a greedy, one-by-one feature elimination strategy, which often overlooks feature coupling effects. In contrast, LeAP jointly optimizes all gates during a single forward pass, effectively capturing global feature importance.

\subsection{Mechanism Analysis: Polarization}

As theoretically analyzed in Section 3.3, LeAP's regularization design explicitly drives gating scores to polarize towards 0 or 1. To verify whether this polarization accurately reflects the genuine predictive utility of features, we conduct a stepwise validation on the MovieLens-1M dataset.

As shown in Table \ref{tab:polarization}, we partition the features into a "Retained Zone" ($g_i > 0.5$) and a "Suppressed Zone" ($g_i \approx 0$) based on the converged gating scores $g_i$. The validation is divided into two phases: first, we incrementally construct the model using features from the Retained Zone (Steps 1-3) to observe performance gains; second, we individually introduce features from the Suppressed Zone into the peak-performance model (Steps 4-9) to assess their redundancy.

\begin{table}[htbp]
\centering
\caption{Stepwise Validation of the LeAP's feature importances (on MovieLens-1M)}
\label{tab:polarization}
\resizebox{\columnwidth}{!}{
\begin{tabular}{l|l|c|c|c|l}
\toprule
\textbf{Step} & \textbf{Feature Set} & \textbf{Gate Score} & \textbf{Val AUC} & $\Delta$ \textbf{AUC} & \textbf{Utility Analysis} \\
\midrule
\multicolumn{6}{l}{\textit{Part 1: Constructing with Retained Zone Features (High Gate > 0.5)}} \\
\midrule
1 & [Title] & 0.995 & 0.7336 & - & Strong predictive signal \\
2 & + User\_ID & 0.941 & 0.8073 & +0.0737 & Strong predictive signal \\
3 & + Genres (Peak Model) & 0.757 & \textcolor{red}{\textbf{0.8105}} & \textcolor{red}{\textbf{+0.0032}} & Weak signal retention \\
\midrule
\multicolumn{6}{l}{\textit{Part 2: Validating Suppressed Zone Features (Base = Peak Model + Single Suppressed Feature)}} \\
\midrule
4 & + Movie\_ID & $\approx 0.0$ & 0.8099 & -0.0006 & Conditional redundancy \\
5 & + Zip & $\approx 0.0$ & 0.8105 & 0.0000 & No significant impact \\
6 & + Age & $\approx 0.0$ & 0.8106 & +0.0001 & No significant impact \\
7 & + Occupation & $\approx 0.0$ & 0.8106 & +0.0001 & No significant impact \\
8 & + Gender & $\approx 0.0$ & 0.8107 & +0.0002 & No significant impact \\
9 & + Timestamp & $\approx 0.0$ & \textbf{0.7938} & \textbf{-0.0167} & Harmful noise \\
\bottomrule
\end{tabular}
}
\end{table}

\noindent\textbf{Results Analysis:}
\begin{itemize}
    \item \textbf{Retention of Weak Signals:} Although the score for \textit{Genres} (0.757) is lower than primary features, its inclusion lifts the AUC by 0.0032. This indicates LeAP effectively identifies and retains weak signals beneficial for generalization, rather than merely filtering out low-weight features.
    \item \textbf{Filtering Conditional Redundancy:} \textit{Movie\_ID} is informative when used alone; however, introducing it when \textit{Title} is already known slightly degrades performance (-0.0006). LeAP assigns it a gate score near 0, accurately capturing the co-linearity and redundancy between these features.
    \item \textbf{Noise Suppression:} Features in Steps 5-8 contribute negligibly to performance; retaining them only inflates the model size. Notably, introducing \textit{Timestamp} causes a significant AUC drop (-0.0167). LeAP successfully truncates such noise features that induce overfitting, proving the reliability of the polarization mechanism in streamlining the feature space.
\end{itemize}

\subsection{Industrial Deployment \& Ablation Study}

In the real-world industrial  system, the search ranking model utilizes a concatenated feature representation exceeding 12,000 dimensions. This massive input tensor incurs significant CPU-GPU communication bandwidth overhead during both training and inference. To reduce dimensionality without compromising predictive performance, we compare the Return on Investment (ROI: pruned dimensions vs. GAUC difference) of different baselines.

\noindent\textbf{Experimental Setup:} Because the industrial model contains over 500 feature fields, traditional Permutation methods require an equivalent number of independent, full-model inferences. Constrained by computational bottlenecks, we evaluate the Permutation method only on a 1\% uniformly sampled subset of the data. In contrast, thanks to their $O(1)$ computational complexity, both LeAP and its ablated version without adaptive regularization (\textbf{LeAP-Base}) complete the feature importance search on the full training dataset.
Remarkably, the total  time for LeAP on the full dataset is even less than that required by the traditional Permutation method to process the 1\% sampled subset.

\begin{table}[htbp]
\centering
\caption{Industrial Feature Pruning ROI and Ablation Analysis (Offline CTR GAUC Diff). $\mathbf{10^{-3}}$ is usually considered as a significant difference for our metric.}
\label{tab:industrial_roi}
\resizebox{\columnwidth}{!}{
\begin{tabular}{l|c|c|c|c}
\toprule
\textbf{Pruned Dims} & \textbf{\%} & \textbf{Permutation}  & \textbf{LeAP-Base} ( w/o Adaptive Reg.) & \textbf{LeAP}  \\
\midrule
$\sim$500 & $\sim$4\% & - $1 \times 10^{-4}$ & No Diff & No Diff \\
$\sim$1900 & $\sim$15\% & - $\mathbf{3 \times 10^{-3}}$ & - $2 \times 10^{-4}$ & No Diff \\
$\sim$2500 & $\sim$20\% & N/A & - $\mathbf{1 \times 10^{-3}}$ & No Diff \\
$\sim$3600 & $\sim$30\% & N/A & - $\mathbf{2 \times 10^{-2}}$ & No Diff \\
$\sim$4300 & $\sim$35\% & N/A & N/A & - $2 \times 10^{-4}$ \\
$\sim$5100 & $\sim$40\% & N/A & N/A & - $3 \times 10^{-4}$ \\
$\sim$6400 & $\sim$50\% & N/A & N/A & - $\mathbf{1 \times 10^{-3}}$ \\
\bottomrule
\end{tabular}
}
\end{table}

\noindent\textbf{Results and Ablation Analysis:}
\begin{itemize}
    \item \textbf{Impact of Training Efficiency and Data Scale:} Comparing Permutation with LeAP-Base, the former suffers a performance drop of $3 \times 10^{-3}$ when pruning 1,900 dimensions due to severe downsampling. Conversely, LeAP-Base, leveraging joint optimization on the full dataset, experiences a negligible drop of only $2 \times 10^{-4}$ at the same pruning ratio. This highlights the substantial engineering advantage of end-to-end learning mechanisms on massive data scales.
    \item \textbf{Necessity of Adaptive Regularization (Ablation):} Table \ref{tab:industrial_roi} shows that LeAP-Base (without adaptive regularization) suffers a severe GAUC degradation of $2 \times 10^{-2}$ when pruning over 3,600 dimensions. This confirms that uniform regularization heavily over-penalizes high-dimensional and sparse features in heterogeneous spaces. In contrast, the complete \textbf{LeAP} maintains consistent performance at the same pruning ratio. Even under an aggressive setting of pruning 50\% of the dimensions, the performance drop is strictly bounded to around $1 \times 10^{-3}$, demonstrating vastly superior robustness.
\end{itemize}

\noindent\textbf{Online A/B Testing:}
Based on the offline ROI evaluation, we deployed the strategy of pruning over 3,600 feature dimensions to the online serving system. After removing the corresponding network structures and performing lightweight fine-tuning, multi-week online A/B testing revealed that LeAP successfully alleviated memory bandwidth pressure while maintaining completely stable (\textit{Neutral}) core business metrics. This  validates the tremendous practical utility of LeAP in ultra-large-scale, complex industrial applications.

\section{Conclusion}

This paper presents LeAP, a plug-in module that transforms the inefficient random permutation process into a learnable mechanism. By introducing Adaptive Regularization based on permutation divergence, LeAP successfully aligns the penalty scale with the task gradient, effectively resolving long-standing evaluation biases in heterogeneous and sparse industrial feature spaces.

Experimental results on four public benchmarks demonstrate that LeAP consistently achieves state-of-the-art performance. Furthermore, its deployment in a billion-requests production system proves its practical efficacy: LeAP successfully pruned over 30\% (3,600+) of input dimensions with zero degradation in business metrics. LeAP offers a scalable and interpretable solution for optimizing next-generation, massive-scale recommender systems.

\begin{credits}
\subsubsection{\ackname} Special thanks go to Jiayu Feng for her support and for crafting the schematic diagrams of the LeAP module.
\end{credits}
% \begin{credits}

% \subsubsection{\discintname}
% \textbf{The authors have no competing interests to declare that are
% relevant to the content of this article.}

% \end{credits}
%
% ---- Bibliography ----
%
% BibTeX users should specify bibliography style 'splncs04'.
% References will then be sorted and formatted in the correct style.
%
\bibliographystyle{splncs04}
\bibliography{mybibliography}
%% Note that this preceding line implies that you store your BibTeX references in a file called 'mybibliography.bib'. If you instead store your references in a file with a different name, for instance 'references.bib', the preceding line should read '\bibliography{references}'. Whatever you do, DO NOT put the file name extension .bib inside the \bibliography command; this will trip up LaTeX compilers. 
% %
% % If you do not want to use BibTeX, you can also type up the bibliography exactly as you see fit, using the following structure:
% \begin{thebibliography}{8}
% % Note that this number 8 reserves an amount of space (equal to the natural width of the given number) for the label of your references; if you have more than 9 references, you will want to change this number to 18. If you have more than 19 references, this number is best changed to 88. If you have more than 99 references, I salute you.
% \bibitem{ref_article1}
% Author, F.: Article title. Journal \textbf{2}(5), 99--110 (2016)

% \bibitem{ref_lncs1}
% Author, F., Author, S.: Title of a proceedings paper. In: Editor,
% F., Editor, S. (eds.) CONFERENCE 2016, LNCS, vol. 9999, pp. 1--13.
% Springer, Heidelberg (2016). \doi{10.10007/1234567890}

% \bibitem{ref_book1}
% Author, F., Author, S., Author, T.: Book title. 2nd edn. Publisher,
% Location (1999)

% \bibitem{ref_proc1}
% Author, A.-B.: Contribution title. In: 9th International Proceedings
% on Proceedings, pp. 1--2. Publisher, Location (2010)

% \bibitem{ref_url1}
% LNCS Homepage, \url{http://www.springer.com/lncs}, last accessed 2023/10/25
% \end{thebibliography}
\end{document}